\documentclass[a4,10pt]{article}

\usepackage{fancybox, enumerate}
\usepackage[dvips]{graphicx, color}
\usepackage{amsmath,amssymb,amsthm, bm, statex}
\usepackage{my-style}
\allowdisplaybreaks[4]

\begin{document}
\begin{center}
{\LARGE{\textsc{Bayesian Reconstruction of Missing Observations}}}\\

\ \\
\ \\
{\Large{Shun Kataoka$^*$, Muneki Yasuda$^\dagger$\footnote{Corresponding author: muneki@yz.yamagata-u.ac.jp}, and Kazuyuki Tanaka$^*$}}\\
\ \\
$^*$Graduate School of Information Sciences, Tohoku University, Sendai 980-8579, Japan\\
$^\dagger$Graduate School of Science and Engineering, Yamagata University, Yonezawa 992-8510, Japan
\end{center}

\subsubsection*{abstract:}
We focus on an interpolation method referred to Bayesian reconstruction in this paper. 
Whereas in standard interpolation methods missing data are interpolated deterministically, 
in Bayesian reconstruction, missing data are interpolated probabilistically using a Bayesian treatment.
In this paper, we address the framework of Bayesian reconstruction and its application to the traffic data reconstruction problem in the field of traffic engineering. 
In the latter part of this paper, we describe the evaluation of the statistical performance of our Bayesian traffic reconstruction model using a statistical mechanical approach
and clarify its statistical behavior.

\section{Introduction}

\begin{figure}[hbt]
\begin{center}
\includegraphics[height=2.5cm]{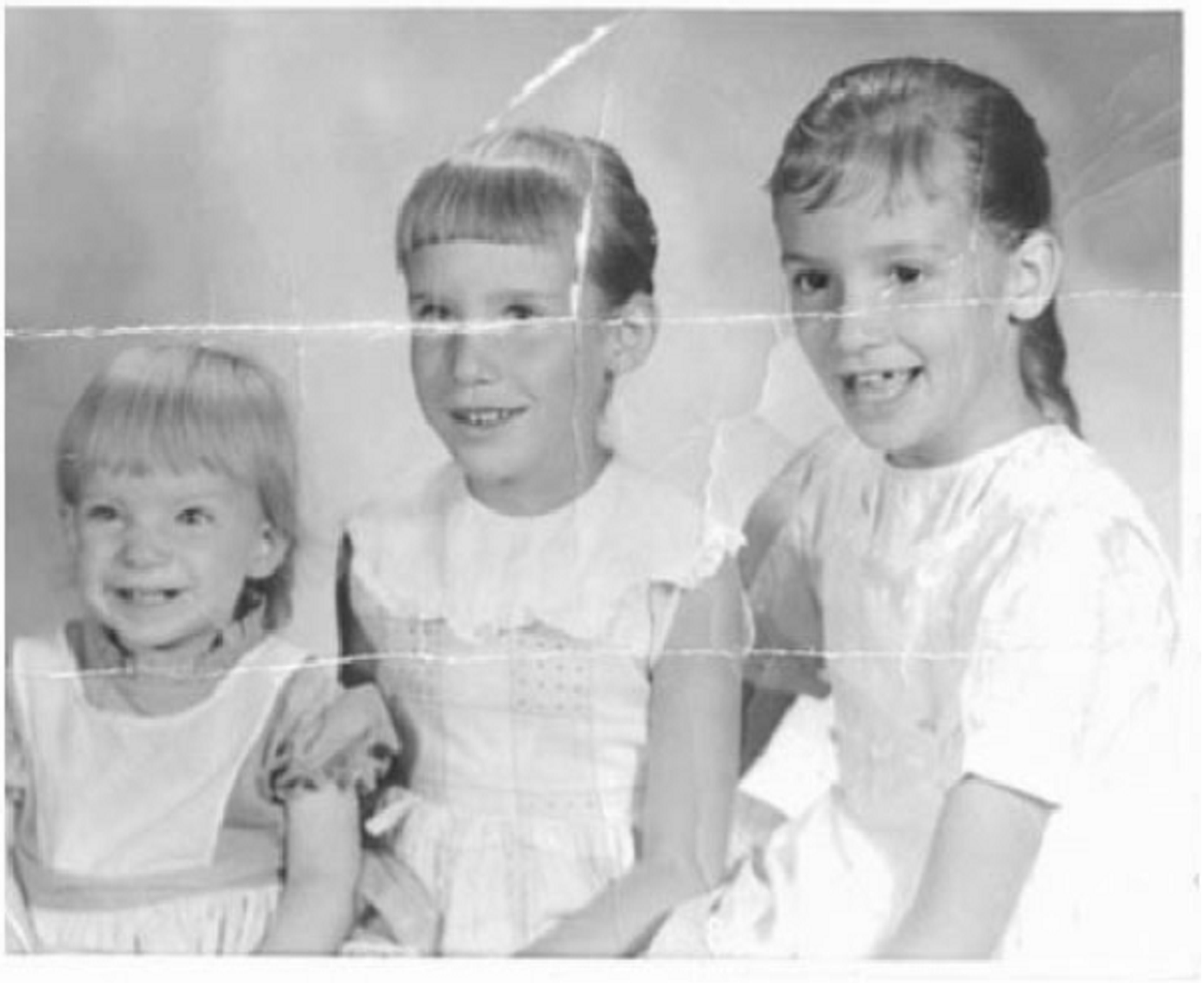}
\includegraphics[height=2.5cm]{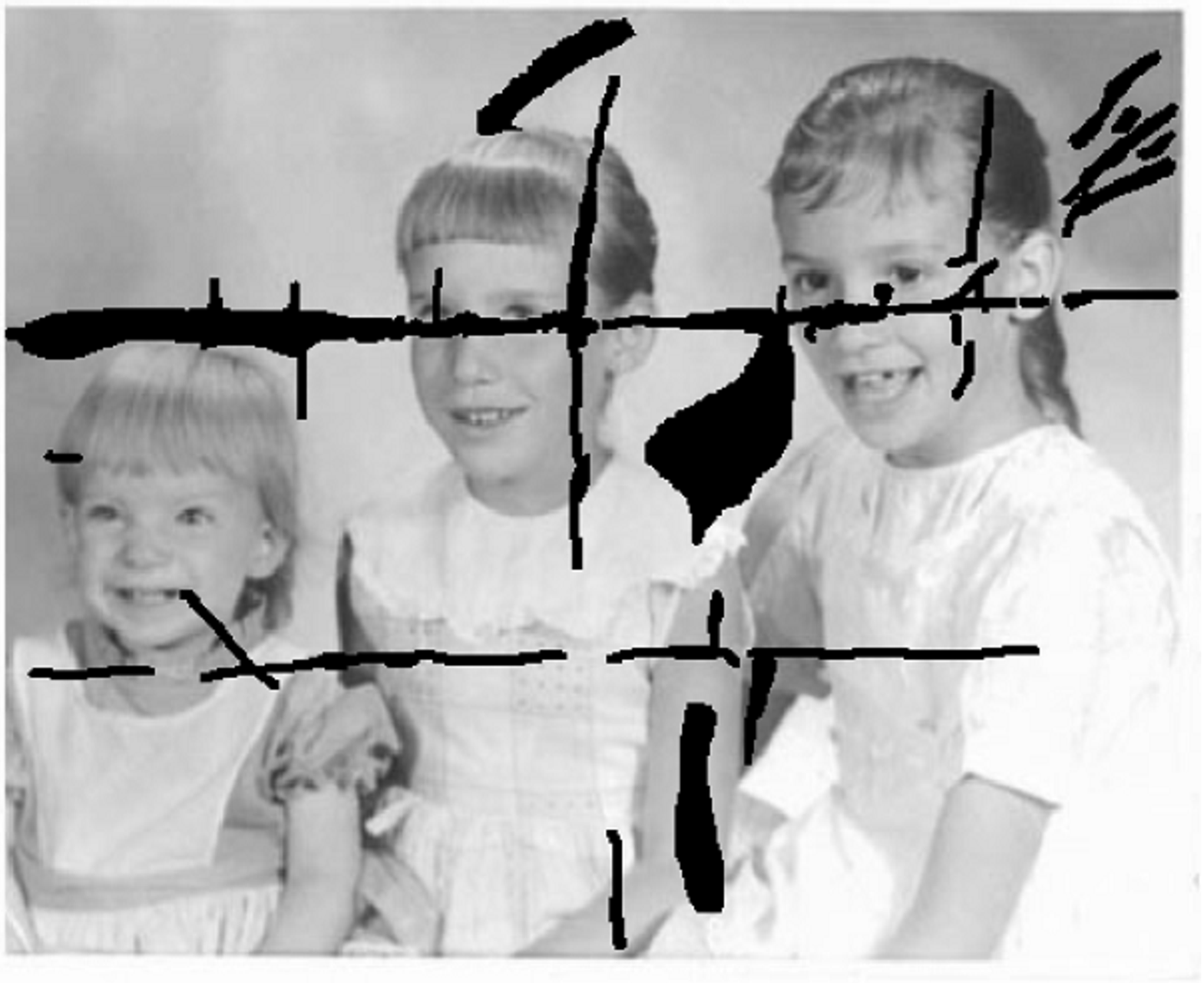}  
\includegraphics[height=2.5cm]{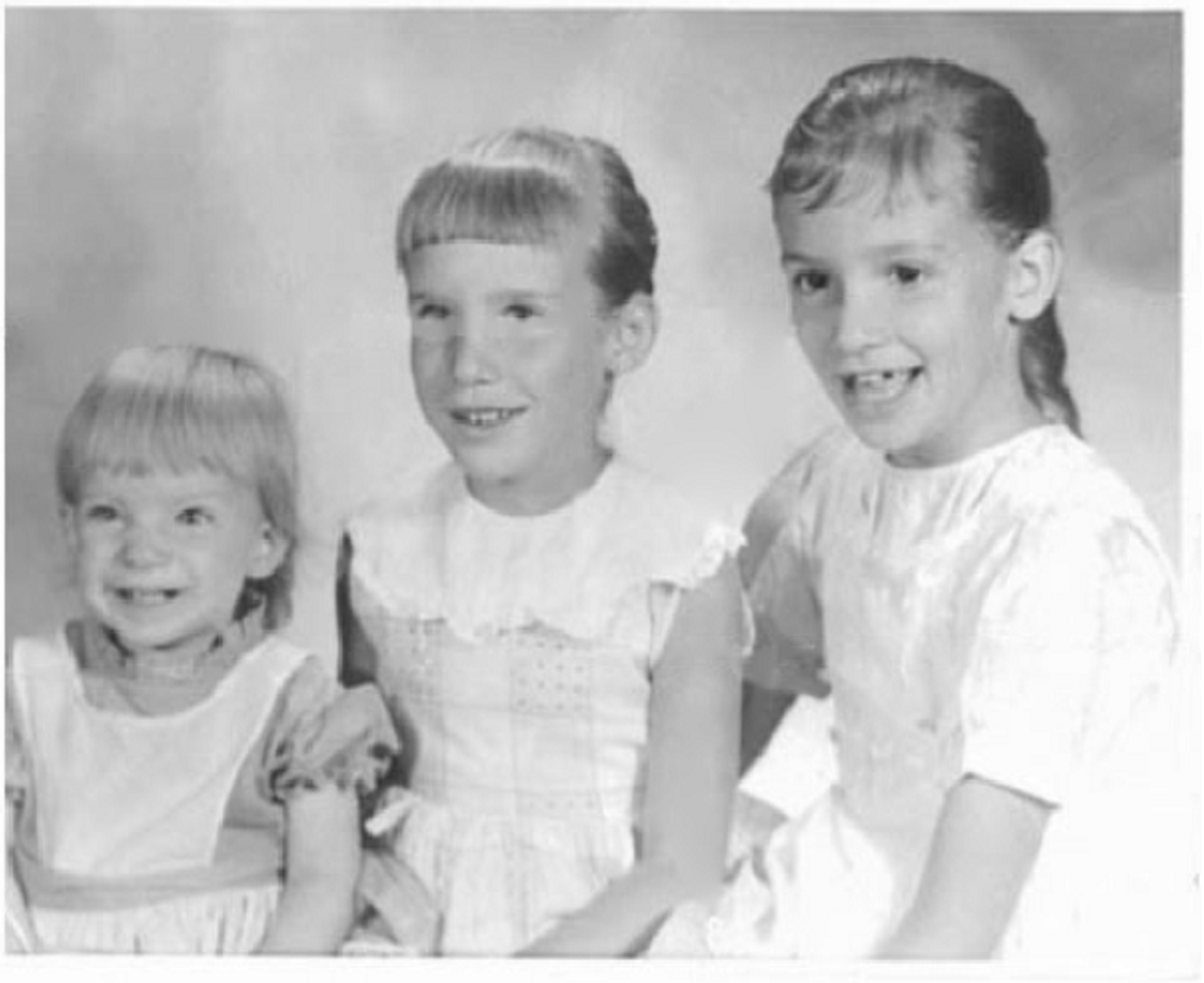}
\end{center}
\caption{
Example of digital image inpainting filter. The left image is the original scratched image and the center image is masked by the black regions. 
We reconstructed the masked region using the digital image inpainting filter to restore the original damaged image. 
The right image is the reconstructed image obtained by using the digital image inpainting method proposed in reference~\cite{YT2013}.
}
\label{fig:inpainting-3ch}
\end{figure}

Methods for interpolating missing data are important in various scientific fields. 
A standard interpolation method, such as spline interpolation, is a deterministic interpolation technique.
An alternative, probabilistic, interpolation technique has been developed in the last few years. 
In the probabilistic interpolation technique, which is called Bayesian reconstruction, 
the Bayesian treatment is used to interpolate and reconstruct missing regions. 
To the best of our Knowledge, Bayesian reconstruction was first implemented in the digital image inpainting filter. 
The digital image inpainting filter is used in the process of reconstructing lost or deteriorated parts of images~\cite{inpainting2000} (see figure \ref{fig:inpainting-3ch}). 
Previously, two of the authors applied Bayesian reconstruction to the digital image inpainting filter~\cite{YOT2005, YOT2006}.
Bayesian reconstruction is now becoming the standard technique in the digital image inpainting filter~\cite{FoE2009}.  

It can be expected that the framework of Bayesian reconstructions will be utilized in various reconstruction problems,   
and therefore, their use should not be limited to image processing. 
Recently, the authors applied Bayesian reconstruction to the traffic data reconstruction problem~\cite{KYT2014}. 
Traffic data reconstruction is an important processing that precedes traffic prediction, such as travel time prediction, density prediction, and route planning. 
In order to provide accurate information to drivers, a broad-scale database of real-time vehicular traffic over an entire city is required. 
However, in practice, it is difficult to collect the traffic data for an entire city, because traffic sensors are not installed on all roads. 
Therefore, the objective of the traffic data reconstruction is to reconstruct the states of unobserved roads where traffic sensors are not installed 
by using information from observed roads where traffic sensors are installed. 

In the first part of this paper, we introduce the details of Bayesian reconstruction, 
and subsequently, an overview of the Bayesian traffic data reconstruction method proposed in reference \cite{KYT2014}, together with some new numerical results. 
In the latter part of this paper, we show a statistical mechanical analysis of our Bayesian traffic reconstruction, 
and clarify its statistical performance.
The remainder of this paper is organized as follows.
In section \ref{sec:BayesianReconst}, we introduce the framework of Bayesian reconstructions based on Markov random fields (MRFs). 
We explain a machine learning strategy for model selection based on the maximum likelihood estimation (MLE) in section \ref{sec:MLE}. 
In section \ref{sec:BayesianTrafficReconst}, we present an overview of Bayesian traffic data reconstruction according to the method proposed in reference \cite{KYT2014}, 
and we show some new numerical results in section \ref{sec:Numerical-Reconstruction}.
We describe our evaluation of the statistical performance of our Bayesian traffic reconstruction in terms of a statistical mechanical analysis in section \ref{sec:MF-Analysis}. 
Finally, we present the conclusions of this paper and outline future work in section \ref{sec:conclusion}.

\section{Scheme of Bayesian Reconstruction of Missing Observations} \label{sec:BayesianReconst}

In the Bayesian framework, we suppose observations (observed data) are probabilistically drawn from a specific probability distribution, 
referred to as prior probability, 
because observations suffer from uncertainty, whose origin is physical noise, incompleteness of some elements, and so on.

Suppose that there exists an $n$-dimensional observation $\bm{x} = \{x_i \in \mathbb{R}\mid i \in V =\{1,2,\ldots,n\}\}$, 
which is generated from prior probability $P_{\mrm{prior}}(\bm{x})$, 
and that we cannot observe a part of the elements in the observation for some reason. 
Since the observation is probabilistically generated, we can treat its elements as random variables.
We define the set of labels of missing elements by $\mcal{M}\subseteq V$ and the complementary set of $\mcal{M}$ by notation $\mcal{O}$, 
i.e., $\mcal{O}:=V\setminus \mcal{M}$, and therefore, $\mcal{O}$ is the set of labels of observed elements.
Given an observation $\bm{x}$, we describe the values of the observed elements by notation $y$ to distinguish them from unobserved elements 
and collectively express the observed elements by $\bm{y} =\{y_i \in \mathbb{R} \mid i \in \mcal{O}\}$. 
The values of the elements in set $\mcal{O}$ are fixed by the observation $\bm{y}$.
Bayesian reconstruction considered in this paper consists of reconstructing the unobserved elements, $\mcal{M}$, in the observation 
by using the observed elements, $\mcal{O}$; 
in other words, the objective is to estimate the values of $\bm{x}_{\mcal{M}}$ by using $\bm{y}$, 
where notation $\bm{x}_{\mcal{A}}$ is the set of $x_i$ belonging to set $\mcal{A} \subseteq  V$, i.e., $\bm{x}_{\mcal{A}}:=\{x_i \mid i \in \mcal{A}\}$.

In order to reconstruct the missing elements in terms of the Bayesian point of view, 
we first formulate the posterior probability of the missing elements, $P(\bm{x}_{\mcal{M}} \mid \bm{y})$, by the Bayesian rule
\begin{align}
P(\bm{x}_{\mcal{M}} \mid \bm{y}) =\frac{P(\bm{x}_{\mcal{M}}, \bm{y})}{P(\bm{y})},
\label{eq:posterior-0}
\end{align}
where the value of $\bm{y}$ is fixed by the observation.
By using Dirac's delta, we have
\begin{align}
P(\bm{x}_{\mcal{M}}, \bm{y})=\Big(\prod_{i \in \mcal{O}}\delta(y_i - x_i)\Big) P(\bm{x}_{\mcal{M}}, \bm{x}_{\mcal{O}})
=\Big(\prod_{i \in \mcal{O}}\delta(y_i - x_i)\Big) P_{\mrm{prior}}(\bm{x}).
\label{eq:posterior-1}
\end{align}
It should be noted that, if $\bm{x}$ are discrete variables, Dirac's delta is replaced by Kronecker's delta.
From equations (\ref{eq:posterior-0}) and (\ref{eq:posterior-1}), we have 
\begin{align}
P(\bm{x}_{\mcal{M}} \mid \bm{y}) \propto P_{\mrm{likelihood}}(\bm{y} \mid \bm{x}) P_{\mrm{prior}}(\bm{x}),
\label{eq:posterior}
\end{align}
where
\begin{align*}
P_{\mrm{likelihood}}(\bm{y} \mid \bm{x}):= \prod_{i \in \mcal{O}}\delta(y_i - x_i).
\end{align*}
Probability $P_{\mrm{likelihood}}(\bm{y} \mid \bm{x})$ is referred to as the likelihood in the Bayesian framework. 
In Bayesian reconstruction, we consider the suitable reconstructed values of unobserved elements, $\bm{x}_{\mcal{M}}^*$, to be the values of $\bm{x}_{\mcal{M}}$ 
that maximize the posterior probability in equation (\ref{eq:posterior}), i.e.,
\begin{align}
\bm{x}_{\mcal{M}}^* = \argmax_{\bm{x}_{\mcal{M}}}P(\bm{x}_{\mcal{M}} \mid \bm{y}) = \argmax_{\bm{x}_{\mcal{M}}}P_{\mrm{likelihood}}(\bm{y} \mid \bm{x}) P_{\mrm{prior}}(\bm{x}).
\label{eq:MAP}
\end{align}
%The schematic illustration of the Bayesian reconstruction is shown in figure \ref{}.

The above reconstruction scheme requires prior probabilities that describe the hidden probabilistic mechanisms of observations. 
However, unfortunately, in almost all situations we do not know the details of the prior probabilities.
Therefore, in order to implement the Bayesian reconstruction system, we should model unknown prior probabilities.

\subsection{Prior Modeling based on Markov Random Fields}

One of the models of prior probabilities of observed data that is presently available is MRF. 
MRFs can easily treat complex spatial interactions among observational data points that create a variety of appearance patterns. 

Consider an undirected graph $G(V,E)$, where $V = \{ 1,2,\ldots,n\}$ is the set of vertices and $E=\{(i,j)\}$ is the set undirected edges between the vertices.  
An MRF is usually defined on such an undirected graph $G(V,E)$ by assigning each variable $x_i$ to the corresponding vertex $i$. 
Edge $(i,j)$ expresses a spatial interaction between variable $x_i$ and variable $x_j$. 
To construct a probabilistic model, we define the joint probability of $\bm{x}$, $P_{\mrm{model}}(\bm{x})$. 
On the undirected graph $G(V,E)$, if we assume a spatial Markov property among random variables, $\bm{x}$, and the positivity of model $P_{\mrm{model}}(\bm{x})>0$, 
by the Hammersley-Clifford theorem, the model can be expressed as
\begin{align}
P_{\mrm{model}}(\bm{x})=\frac{1}{Z}\exp\Big(\sum_{i \in V} \phi_i(x_i) + \sum_{(i,j) \in E}\psi_{ij}(x_i, x_j)\Big),
\label{eq:MRF}
\end{align}
without loss of generalities. 
The first term in the exponent, $\phi_i(x_i)$, is a potential function on vertex $i$ that determines the characteristic of $x_i$, 
and the second term in the exponent, $\psi_{ij}(x_i, x_j)$, is a potential function between vertices $i$ and $j$ that determines the interaction between $x_i$ and $x_j$. 
$\sum_{(i,j) \in E}$ represents the summation running over all edges,  
$Z$ denotes the normalization constant, sometimes referred to as the partition function, defined by
\begin{align*}
Z:=\int_{-\infty}^{\infty}\exp\Big(\sum_{i \in V} \phi_i(x_i) + \sum_{(i,j) \in E}\psi_{ij}(x_i, x_j)\Big) \diff \bm{x}.
\end{align*}
The model in equation (\ref{eq:MRF}) is the MRF that is most frequently used. In the MRF, let us consider the conditional probability of $x_i$ expressed as 
\begin{align}
P_{\mrm{model}}(x_i \mid \bm{x}_{-i})=\frac{P_{\mrm{model}}(\bm{x})}{\int_{-\infty}^{\infty}P_{\mrm{model}}(\bm{x}) \diff x_i},
\label{eq:Conditional_MRF-0}
\end{align}
where $\bm{x}_{-i}$ denotes the set of all variables except $x_i$: $\bm{x}_{-i}=\{x_j \mid j \in V \setminus \{i\}\}$. Equations (\ref{eq:MRF}) and (\ref{eq:Conditional_MRF-0}) lead to
\begin{align}
P_{\mrm{model}}(x_i \mid \bm{x}_{-i}) = P_{\mrm{model}}(x_i \mid \bm{x}_{\partial(i)}),
\label{eq:Conditional_MRF-1}
\end{align}
where $\partial(i)$ denotes the set of vertices connecting to vertex $i$ in the graph, 
and $\bm{x}_{\partial(i)}$ denotes the set of variables on the vertices belonging to $\partial(i)$, that is, 
$\bm{x}_{\partial(i)}$ is the set of nearest neighbor variables of $x_i$: $\bm{x}_{\partial(i)}= \{x_j \mid j \in \partial(i)\}$.
Equation (\ref{eq:Conditional_MRF-1}) states that the variable $x_i$ depends on only nearest neighbor variables in the conditional probability, 
and this constitutes the spatial Markov property of MRF.

\subsection{Model Selection using Parametric Machine Learning} \label{sec:MLE}

In order to implement the MRF in equation (\ref{eq:MRF}), the forms of potential functions should be determined. 
This is one of the most important points in MRF modeling. 
Parametrically, we model the potential functions by certain parametric functions with parameter $\bm{\theta}$,
\begin{align}
P_{\mrm{model}}(\bm{x} \mid \bm{\theta})=\frac{1}{Z(\bm{\theta})}\exp\Big(\sum_{i \in V} \phi_i(x_i\mid \theta_i) + \sum_{(i,j) \in E}\psi_{ij}(x_i, x_j, \mid \theta_{ij})\Big).
\label{eq:MRF_parametric}
\end{align}
Thus, we should find the optimal values of the parameters. 
The standard method for achieving this is provided by the field of machine learning theory described as follows.

The objective of MRF modeling is to model the unknown prior probability of observation $P_{\mrm{prior}}(\bm{x})$. 
Therefore, the optimal values of the parameters, $\bm{\theta}^*$, should minimize some distance between the prior probability and our model $P_{\mrm{model}}(\bm{x} \mid \bm{\theta})$.
The Kullback-Leibler divergence (KLD)
\begin{align}
\mcal{K}(P_0 || P_1):= \int_{-\infty}^{\infty} P_0(\bm{x}) \ln \frac{P_0(\bm{x})}{P_1(\bm{x})}\diff\bm{x}
\label{eq:def-KLD}
\end{align}
is often utilized as a measure of two distinct probabilities, $P_0(\bm{x})$ and $P_1(\bm{x})$. 
The value of KLD is always non-negative and is zero when two probabilities are equivalent. 
Thus, we consider that two distinct probabilities are close to each other when the value of KLD is small.
In terms of KLD, we suppose the optimal values of the parameters are given by minimizing the value of KLD between the prior probability and our model,
\begin{align*}
\bm{\theta}^* = \argmin_{\bm{\theta}}\mcal{K}(P_{\mrm{prior}} || P_{\mrm{model}}).
\end{align*}
However, we cannot perform this minimization because we do not know the prior probability.

Since we do not know the prior probability, 
we suppose instead that we have many complete observations\footnote{
``Complete'' means each observation includes no missing points.
} generated from the prior probability. 
We describe the observations by $\mcal{D}=\{\bm{y}^{(\mu)} \in \mathbb{R}^n \mid \mu = 1,2,\ldots,N\}$, 
and we define the empirical distribution of the $N$ complete observations by
\begin{align}
Q_{\mcal{D}}(\bm{x}):=\frac{1}{N}\sum_{\mu=1}^N \prod_{i \in V} \delta(y_i^{(\mu)} - x_i).
\end{align} 
The empirical distribution is the frequency distribution of the $N$ complete observations.
It should be noted that, if $\bm{x}$ are discrete variables, Dirac's delta is again replaced by Kronecker's delta. 
We suppose the empirical distribution has some important properties of the prior probability 
and that suppose the optimal values of the parameters are approximately obtained by minimizing the value of KLD between the empirical distribution and our model,
\begin{align}
\bm{\theta}^* = \argmin_{\bm{\theta}}\mcal{K}(P_{\mrm{prior}} || P_{\mrm{model}}) 
\approx \argmin_{\bm{\theta}}\mcal{K}(Q_{\mcal{D}} || P_{\mrm{model}}).
\label{eq:minimize-KLD}
\end{align}
This minimization can be perform if we have the complete observations generated from the prior probability. 
Equation (\ref{eq:minimize-KLD}) is rewritten as
\begin{align*}
\bm{\theta}^*\approx \argmin_{\bm{\theta}}\mcal{K}(Q_{\mcal{D}} || P_{\mrm{model}})
=\argmax_{\bm{\theta}}\int_{-\infty}^{\infty} Q_{\mcal{D}}(\bm{x}) \ln P_{\mrm{model}}(\bm{x} \mid \bm{\theta})\diff \bm{x}.
\end{align*}
This corresponds to the MLE in statistics. 
In the above scheme, we assumed there are no missing points in the $N$ observations used in the estimation of the parameters. 
If the observations include missing points, we will use an alternative strategy and apply the expectation and maximization (EM) algorithm. 

From the above arguments, the (parametric) Bayesian reconstruction system is summarized as follows.
Before the reconstructions, we design the potential functions in our MRF model $P_{\mrm{model}}(\bm{x} \mid \bm{\theta})$ and 
estimate the optimal values of the parameters, $\bm{\theta}^*$, in advance by using many complete observations and equation (\ref{eq:minimize-KLD}). 
Then, the reconstruction is approximately performed using the constructed model $P_{\mrm{model}}(\bm{x} \mid \bm{\theta}^*)$ 
instead of the prior probability in equation (\ref{eq:MAP}), i.e.,
\begin{align}
\bm{x}_{\mcal{M}}^* = \argmax_{\bm{x}_{\mcal{M}}}P_{\mrm{likelihood}}(\bm{y} \mid \bm{x}) P_{\mrm{prior}}(\bm{x})
\approx \argmax_{\bm{x}_{\mcal{M}}}P_{\mrm{likelihood}}(\bm{y} \mid \bm{x}) P_{\mrm{model}}(\bm{x} \mid \bm{\theta}^*).
\label{eq:MAP_practical}
\end{align}
Since
\begin{align*}
\argmax_{\bm{x}_{\mcal{M}}}P_{\mrm{likelihood}}(\bm{y} \mid \bm{x}) P_{\mrm{model}}(\bm{x} \mid \bm{\theta}^*)
=\argmax_{\bm{x}_{\mcal{M}}}P_{\mrm{model}}(\bm{x}_{\mcal{M}} \mid \bm{x}_{\mcal{O}} = \bm{y}, \bm{\theta}^*),
\end{align*}
the suitable reconstructed values of unobserved missing points, $\bm{x}_{\mcal{M}}^*$, are the values that maximize
the conditional probability of our model, 
\begin{align*}
P_{\mrm{model}}(\bm{x}_{\mcal{M}} \mid \bm{x}_{\mcal{O}}, \bm{\theta}^*) = \frac{ P_{\mrm{model}}(\bm{x} \mid \bm{\theta}^*)}
{ \int_{-\infty}^{\infty}P_{\mrm{model}}(\bm{x} \mid \bm{\theta}^*) \diff \bm{x}_{\mcal{M}}},
\end{align*} 
with $\bm{x}_{\mcal{O}}$ fixed by the observation $\bm{y}$.

\section{Overview of Bayesian Traffic Data Reconstruction} \label{sec:BayesianTrafficReconst}

In this section, we give an overview of the application of the Bayesian reconstruction scheme presented in the previous section 
to the traffic data recognition problem proposed by the authors~\cite{KYT2014}, together with some new numerical results.

\subsection{MRF model for Bayesian Traffic Data Reconstruction} \label{sec:Traffic-MRF}

\begin{figure}[hbt]
\begin{center}
\includegraphics[height=3.0cm]{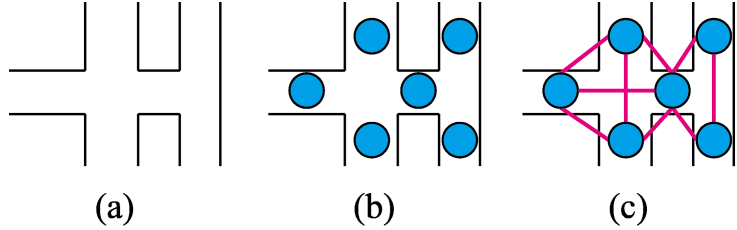} 
\end{center}
\caption{Undirected graph representation for road network. (a) Road network with six roads and two intersections. (b) Vertices are assigned to roads. 
(c) Edges are drawn between two roads that are connected to each other at intersections.}
\label{fig:RoadGraph}
\end{figure}

We applied the Bayesian reconstruction scheme to traffic data reconstruction. 
Our goal is to reconstruct the states of roads, and therefore, random variables $\bm{x}$ are assigned to roads. 
In order to formulate an MRF for a road network, we construct an undirected graph $G(V,E)$ as follows: 
we assign each vertex on each road and draw each edge between two roads that are connected to each other at a traffic intersection (see figure \ref{fig:RoadGraph}). 
On the undirected graph, we define the MRF by
\begin{align}
P_{\mrm{model}}(\bm{x} \mid \bm{\theta})=\frac{1}{Z(\bm{\theta})}\exp\Big(\sum_{i \in V} h_i x_i -\frac{\xi}{2}\sum_{i \in V}x_i^2 
- \frac{J}{2}\sum_{(i,j) \in E}(x_i - x_j)^2\Big),
\label{eq:MRF_traffic}
\end{align}
where $\bm{\theta} = \{ \bm{h}, \xi, J\}$ are the parameters of the model
\footnote{Although this expression seems to differ slightly from the original model proposed in reference \cite{KYT2014}, 
this expression is essentially equivalent to the original model.}.
The variable $x_i \in \mathbb{R}$ expresses the state of road $i$. In this paper, we consider $\bm{x}$ as traffic densities according to the method in reference \cite{KYT2014}. 
The traffic density on road $i$ is defined by the number of cars per unit area on road $i$; 
high densities tend to lead to traffic jams. 
This MRF is obtained by setting $\phi_i(x_i)= h_i x_i - \xi x_i^2 / 2$ and $\psi_{ij}(x_i,x_j)=J(x_i-x_j)^2/2$ in equation (\ref{eq:MRF}). 
Parameter $h_i$ is the bias that controls the level of the traffic density of road $i$, 
and parameter $\xi >0$ controls the variances in the traffic densities.
The interaction term in the last term in the exponent in equation (\ref{eq:MRF_traffic}) corresponds to our assumption 
that is traffic densities of neighboring roads take close values. Parameter $J \geq 0$ controls the strength of the assumption. 
This MRF forms the multi-dimensional Gaussian and is known as the Gaussian graphical model (GGM). 

As in the previous section, we represent the set of unobserved roads by $\mcal{M}$ and the set of observed roads by $\mcal{O}$.  
After determining the values of the parameters by the machine learning method in equation (\ref{eq:minimize-KLD}), 
from equation (\ref{eq:MAP_practical}), the reconstructed densities on the unobserved roads are obtained by
\begin{align}
\bm{x}_{\mcal{M}}^* = \argmax_{\bm{x}_{\mcal{M}}}P_{\mrm{model}}(\bm{x}_{\mcal{M}} \mid \bm{x}_{\mcal{O}} = \bm{y}, \bm{\theta}^*), 
\label{eq:MAP_practical_GGM-0}
\end{align}
where $\bm{y}$ represents the densities on the observed roads.
Since our model in equation (\ref{eq:MRF_traffic}) is multi-dimensional Gaussian, the conditional probability is also multi-dimensional Gaussian. 
Therefore, equation (\ref{eq:MAP_practical_GGM-0}) is rewritten as
\begin{align}
\bm{x}_{\mcal{M}}^* = \int_{-\infty}^{\infty} \bm{x}_{\mcal{M}}P_{\mrm{model}}(\bm{x}_{\mcal{M}} \mid \bm{x}_{\mcal{O}} = \bm{y}, \bm{\theta}^*) \diff \bm{x}_{\mcal{M}}.
\label{eq:MAP_practical_GGM}
\end{align}
Hence, we find that the reconstructed densities, $\bm{x}_{\mcal{M}}^*$, 
are the expectations of $P_{\mrm{model}}(\bm{x}_{\mcal{M}} \mid \bm{x}_{\mcal{O}} = \bm{y}, \bm{\theta}^*)$. 
The expectations are obtained by solving the simultaneous equations, 
\begin{align}
x_i = \frac{1}{\xi + |\partial(i)|J}\Big( h_i + J \sum_{j \in \partial(i)}z_j\Big)
\quad i \in \mcal{M}
\label{eq:MFE}
\end{align}
by an iteration method, where $|\mcal{A}|$ denotes the number of elements in the assigned set $\mcal{A}$ and 
\begin{align*}
z_j=
\begin{cases}
x_j & j \in \mcal{M} \\
y_j & j \in \mcal{O}
\end{cases}
.
\end{align*}
Equation (\ref{eq:MFE}) is known as the mean-field equation, 
which is obtained by the naive mean-field approximation for $P_{\mrm{model}}(\bm{x}_{\mcal{M}} \mid \bm{x}_{\mcal{O}} = \bm{y}, \bm{\theta}^*)$, 
and is also known as the Gauss-Seidel method. 
It is known that in GGM mean-field equations and Gauss-Seidel methods are equivalent in general 
and that mean-field equations always provide exact expectations~\cite{Weiss&Freeman2007}.

\subsection{Results of Numerical Simulation using Road Network of Sendai-city} \label{sec:Numerical-Reconstruction}

\begin{figure}[hbt]
\begin{center}
\includegraphics[height=3.5cm]{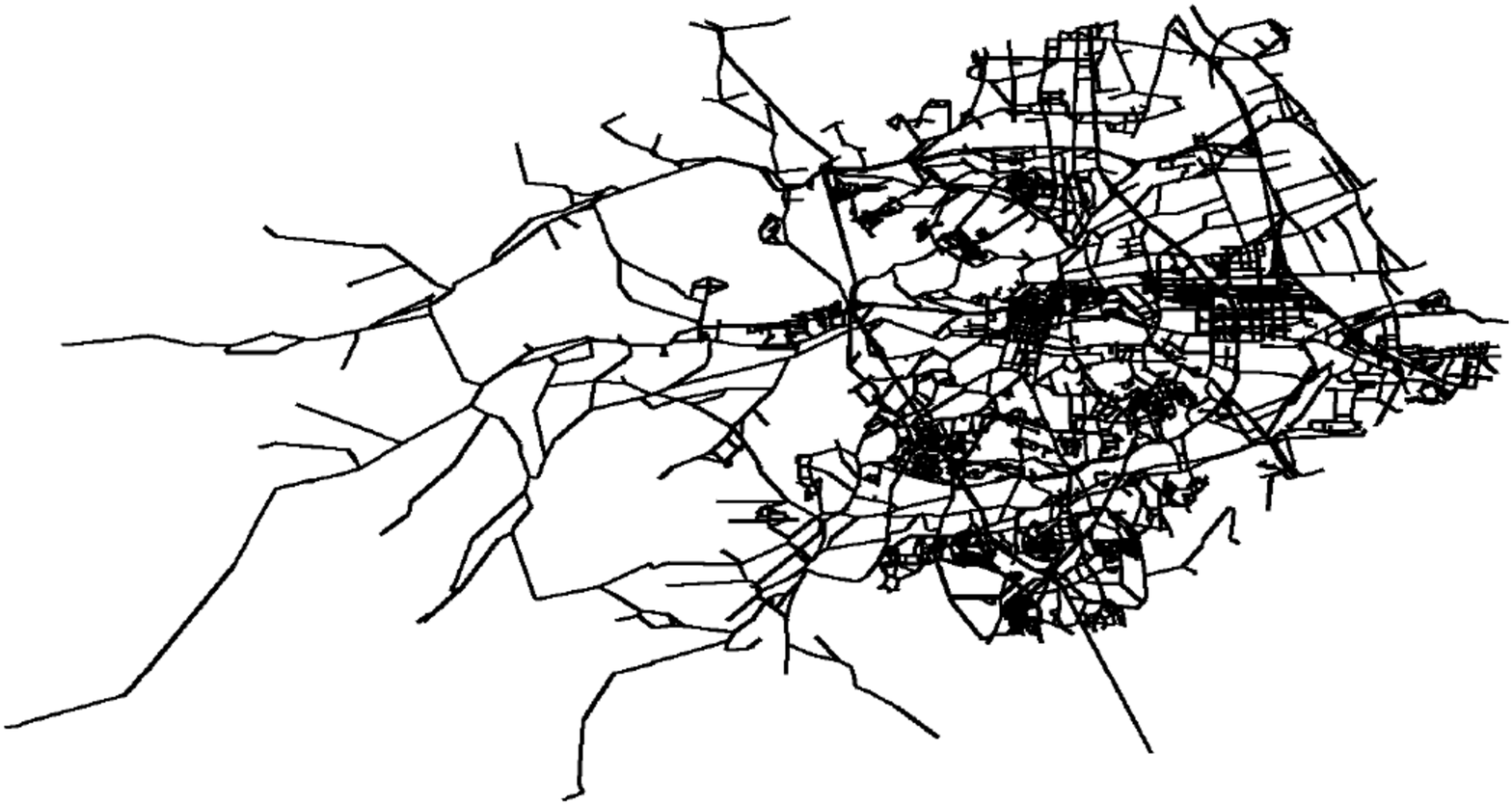} 
\end{center}
\caption{Road network of Sendai, Japan. This network consists of about ten thousand roads.}
\label{fig:sendai}
\end{figure}

In this section, we describe the application of our Bayesian traffic data reconstruction to the road network of the city of Sendai (shown in figure \ref{fig:sendai})
and show the performance of our model. 
Figure \ref{fig:sendai} shows the road network of Sendai which consists of about ten thousand roads. 
According to our Bayesian traffic data reconstruction scheme, first, we defined the MRF model shown in equation (\ref{eq:MRF_traffic}) for the road network. 
The structure of the MRF was constructed according to figure \ref{fig:RoadGraph}. 

The parameters were determined by the maximum likelihood estimation shown in section \ref{sec:MLE} with the $L_2$ regularizations~\cite{KYT2014}. 
Regularizations are frequently used to avoid over-fitting to noises in training data. 
In the maximum likelihood estimation, we used $N = 359$ complete traffic data generated by a traffic simulator for the road network of Sendai. 
Although the traffic data were not real, they were presumed to represent typical behavior of traffic in Sendai. 
Using the MRF model, we reconstructed the traffic densities of Sendai.
\begin{figure}[hbt]
\begin{center}
\includegraphics[height=3.5cm]{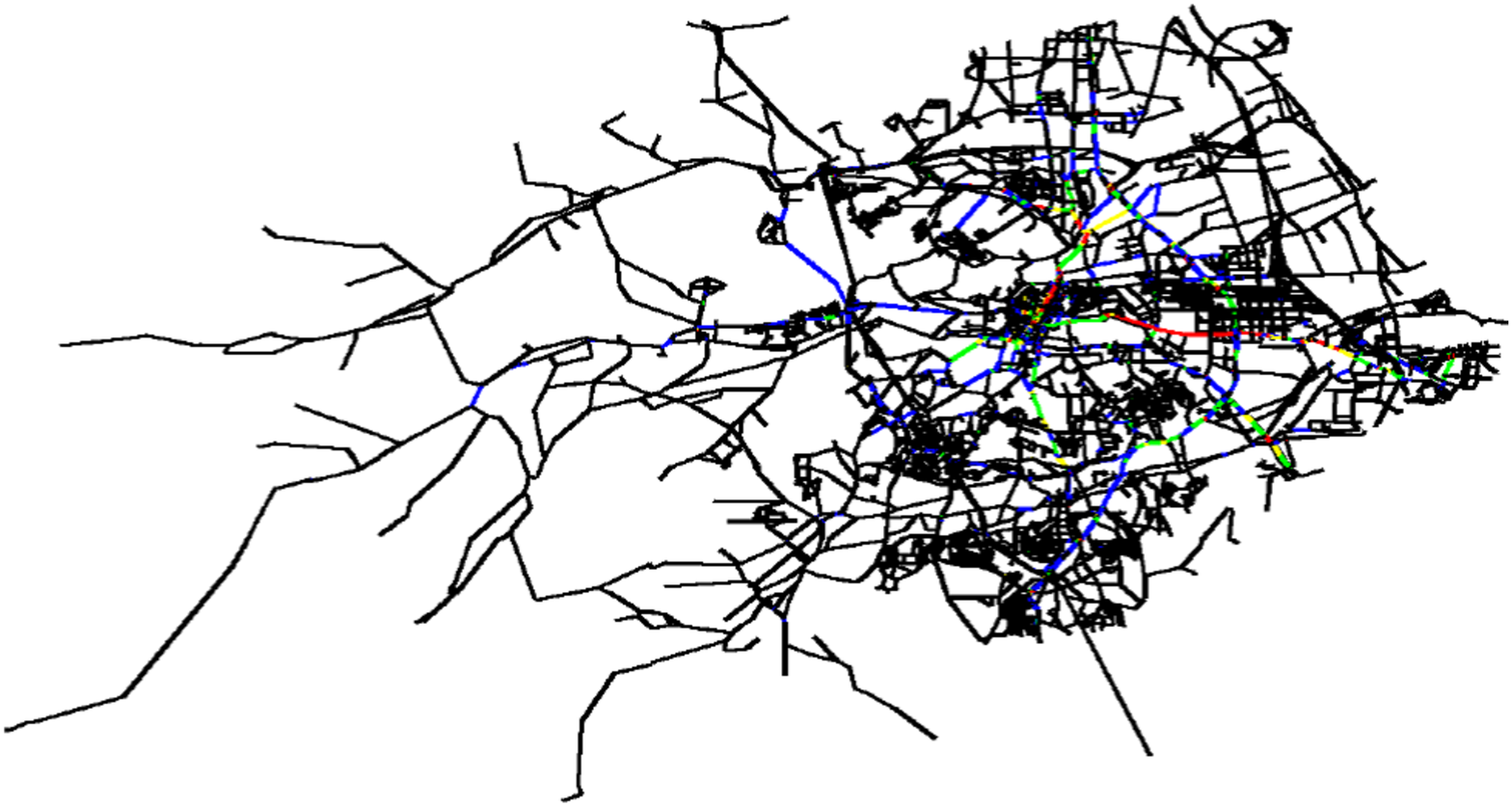} \hspace{4mm}
\includegraphics[height=3.0cm]{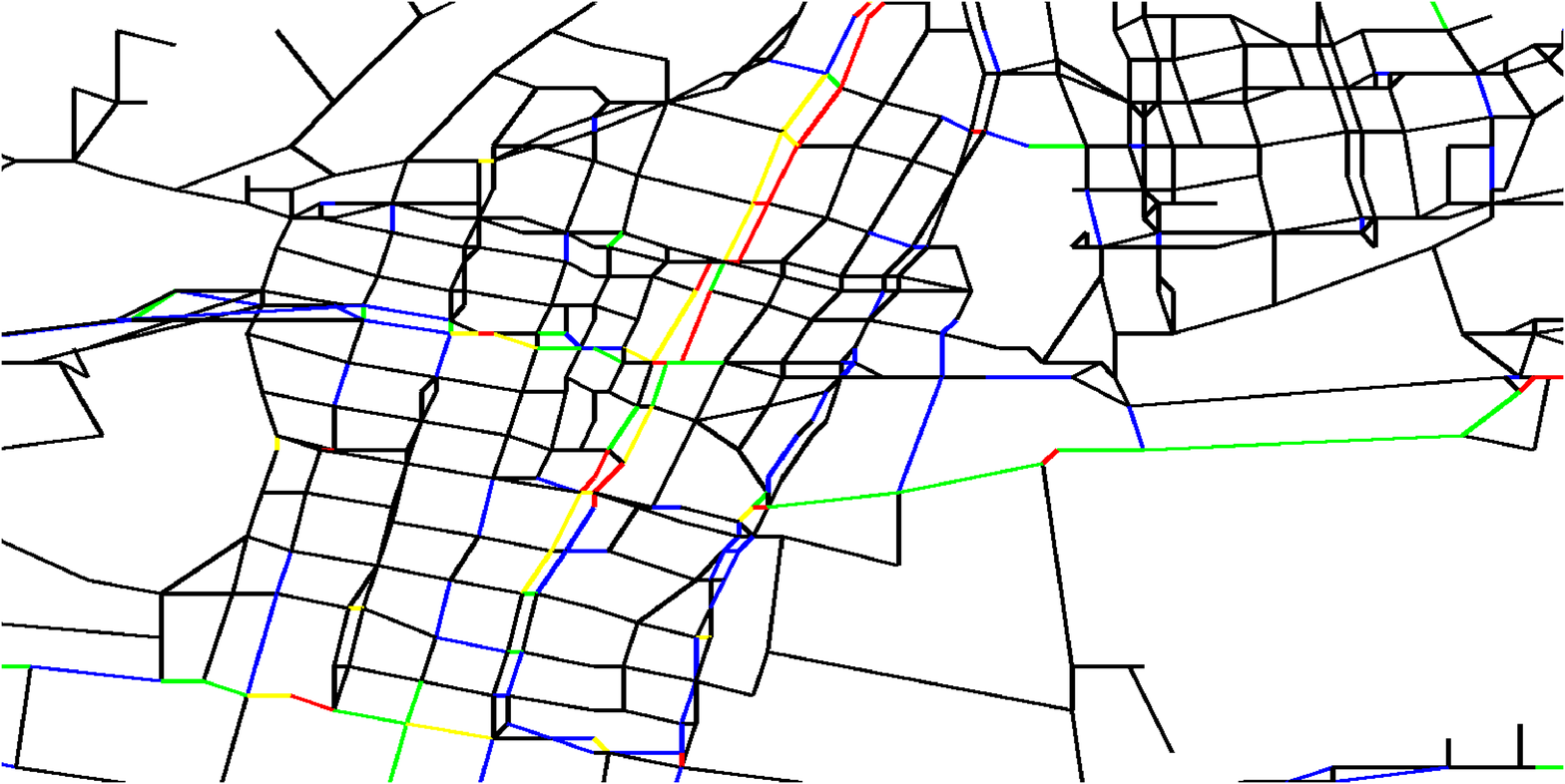}
\end{center}
\caption{True traffic density used in our numerical experiment. Each road is colored according to its traffic density. The network in the left panel is the entire network and the network in the right panel is an enlarged image of the center of Sendai.}
\label{fig:data}
\end{figure}
Figure \ref{fig:data} shows the traffic density data that were not used in the learning, namely, the test data. 
In order to visually represent the densities, we quantized the densities into five stages at 0.03 intervals; 
each road is colored according to its quantized traffic density. 
The densities increase in the following order: black, blue, green, yellow, and red.
Thus, a road colored black has a density in the interval $[0,0.03]$.

\begin{figure}[hbt]
\begin{center}
\includegraphics[height=3.5cm]{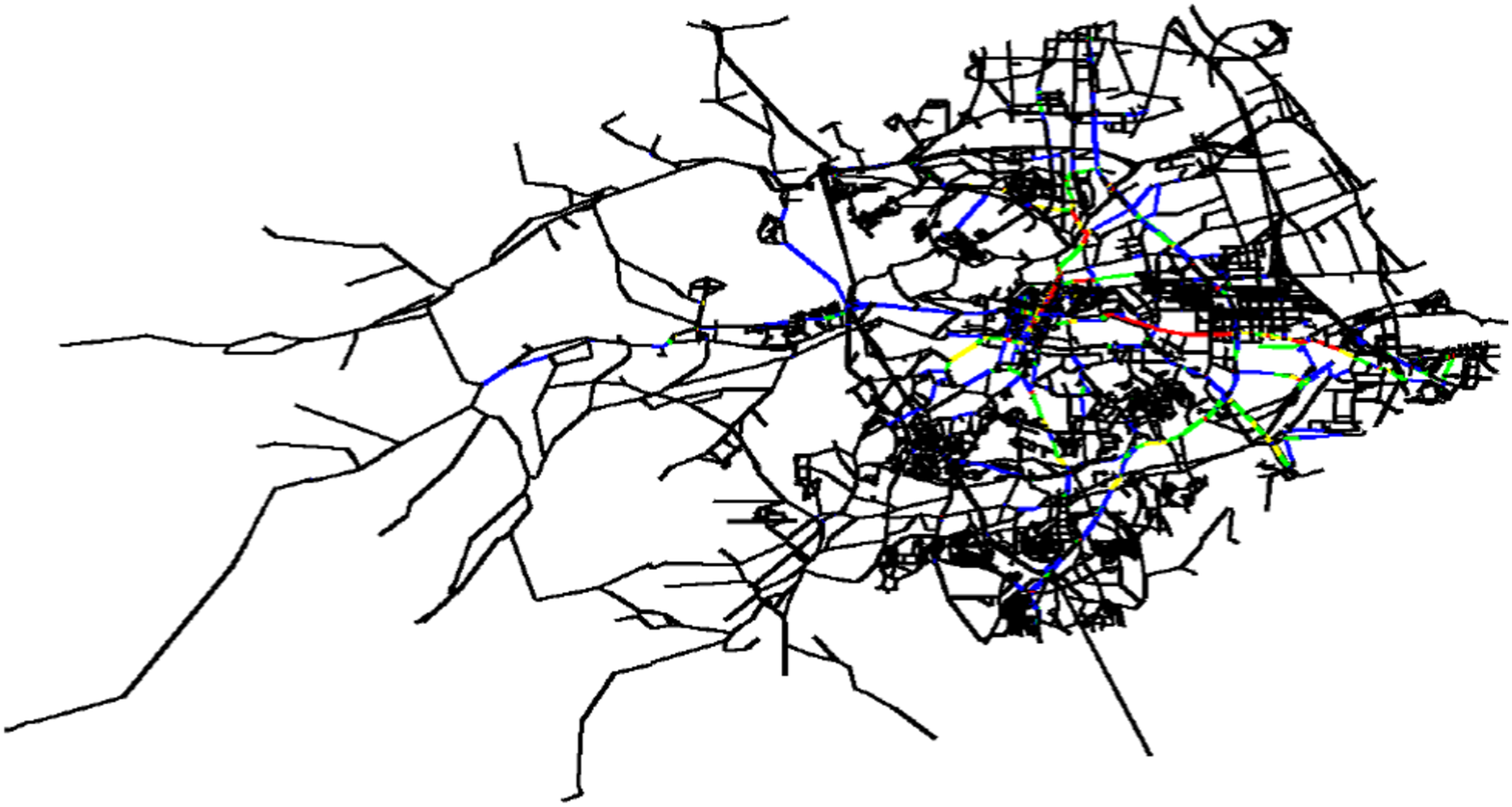} \hspace{4mm}
\includegraphics[height=3.0cm]{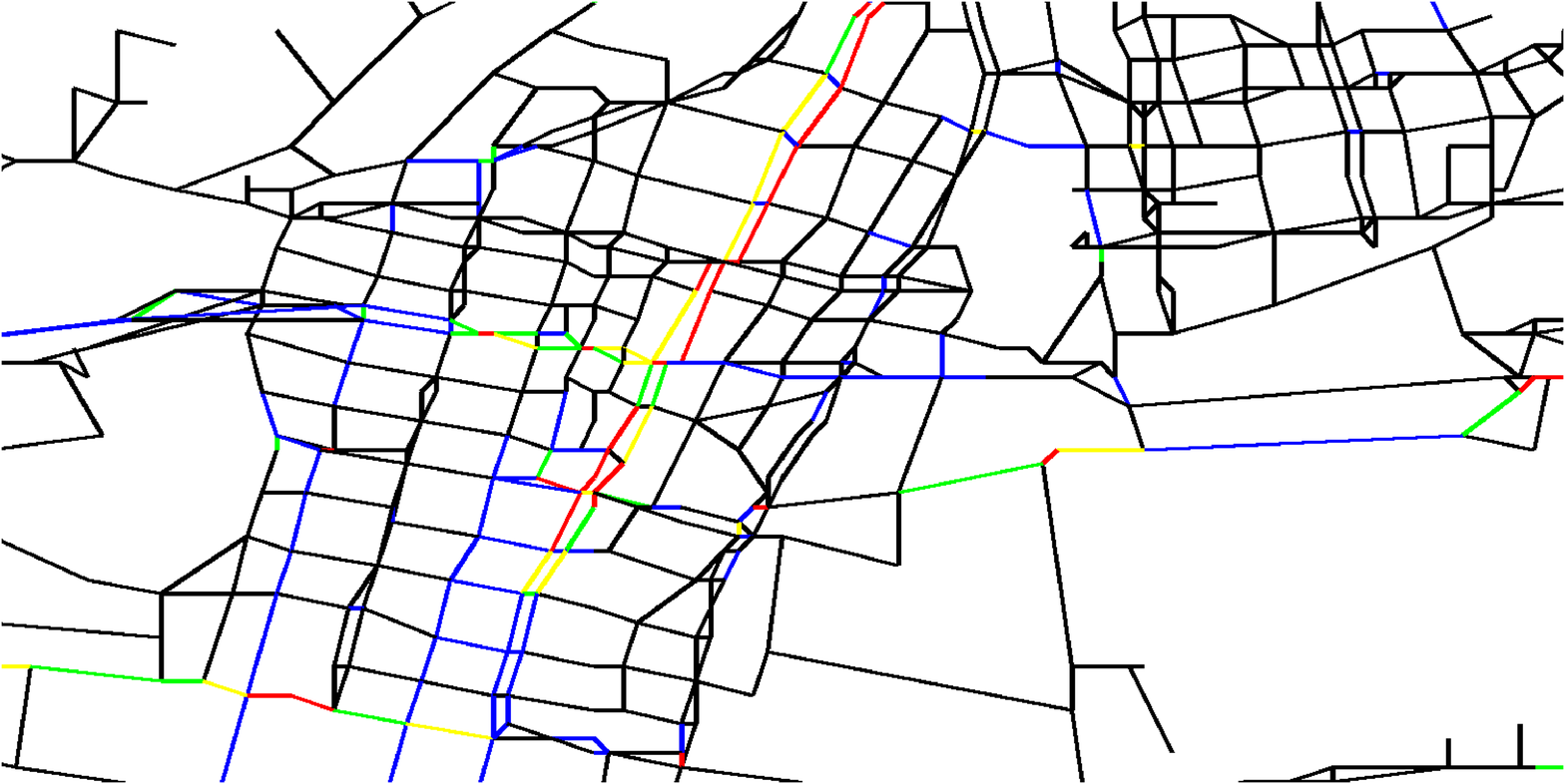} 
\end{center}
\caption{Reconstruction result by using our Bayesian reconstruction, where each road is colored according to its traffic density. 
The network in the left panel is the entire network and the network in the right panel is an enlarged image of the central area of Sendai.}
\label{fig:inf}
\end{figure}

\begin{figure}[hbt]
\begin{center}
\includegraphics[height=4.0cm]{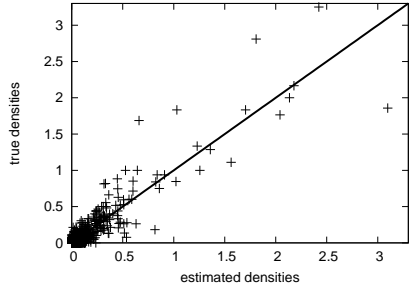}
\end{center}
\caption{Scatter plot of the true densities in figure \ref{fig:data} and the reconstructed densities in figure \ref{fig:inf}. }
\label{fig:scatter}
\end{figure}

For the traffic density data shown in figure \ref{fig:data}, we suppose that the densities in some roads are unobserved and  
we randomly select unobserved roads with probability $p = 0.8$. 
\begin{figure}[hbt]
\begin{center}
\includegraphics[height=3.5cm]{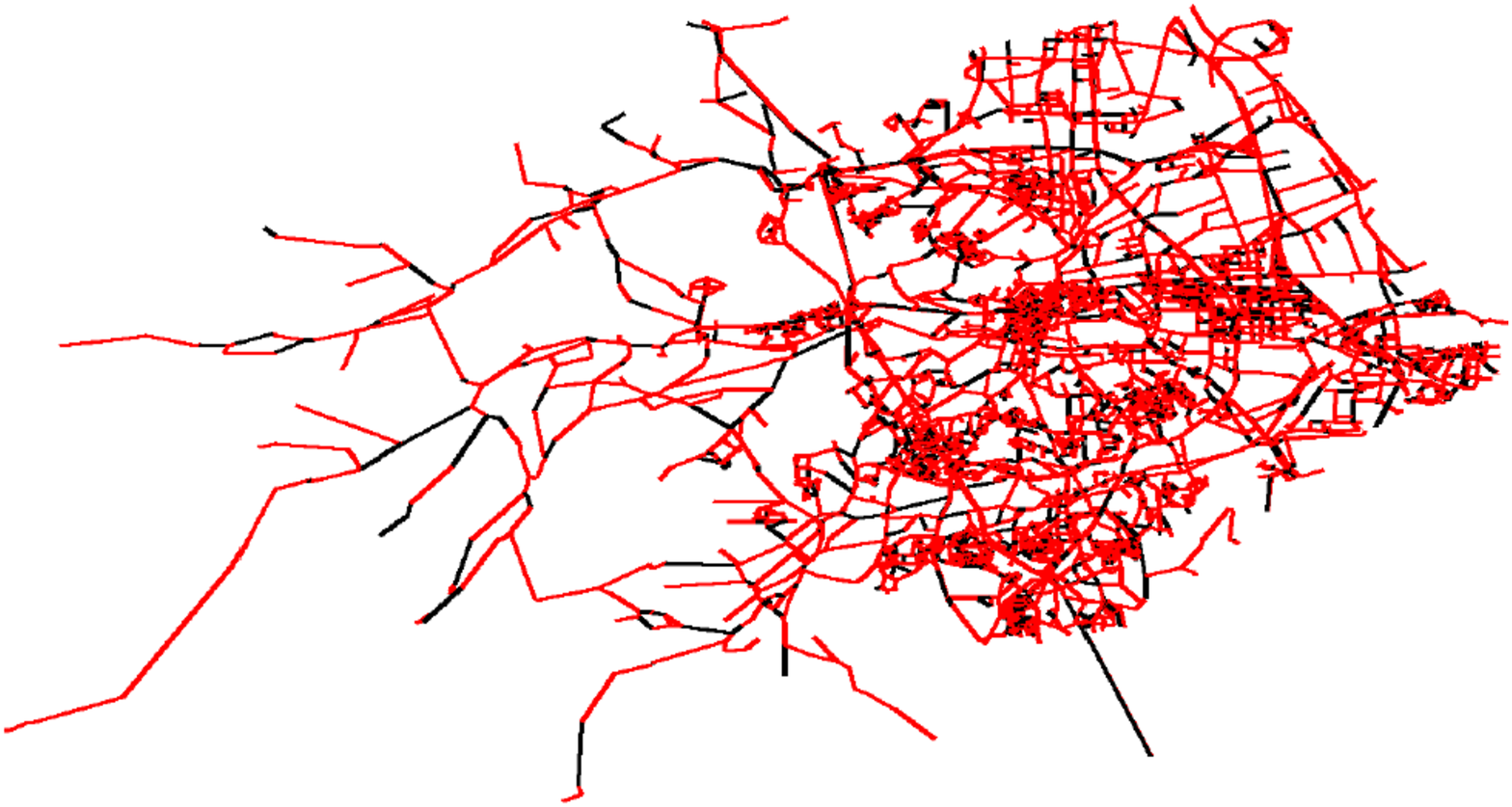} \hspace{4mm}
\includegraphics[height=3.0cm]{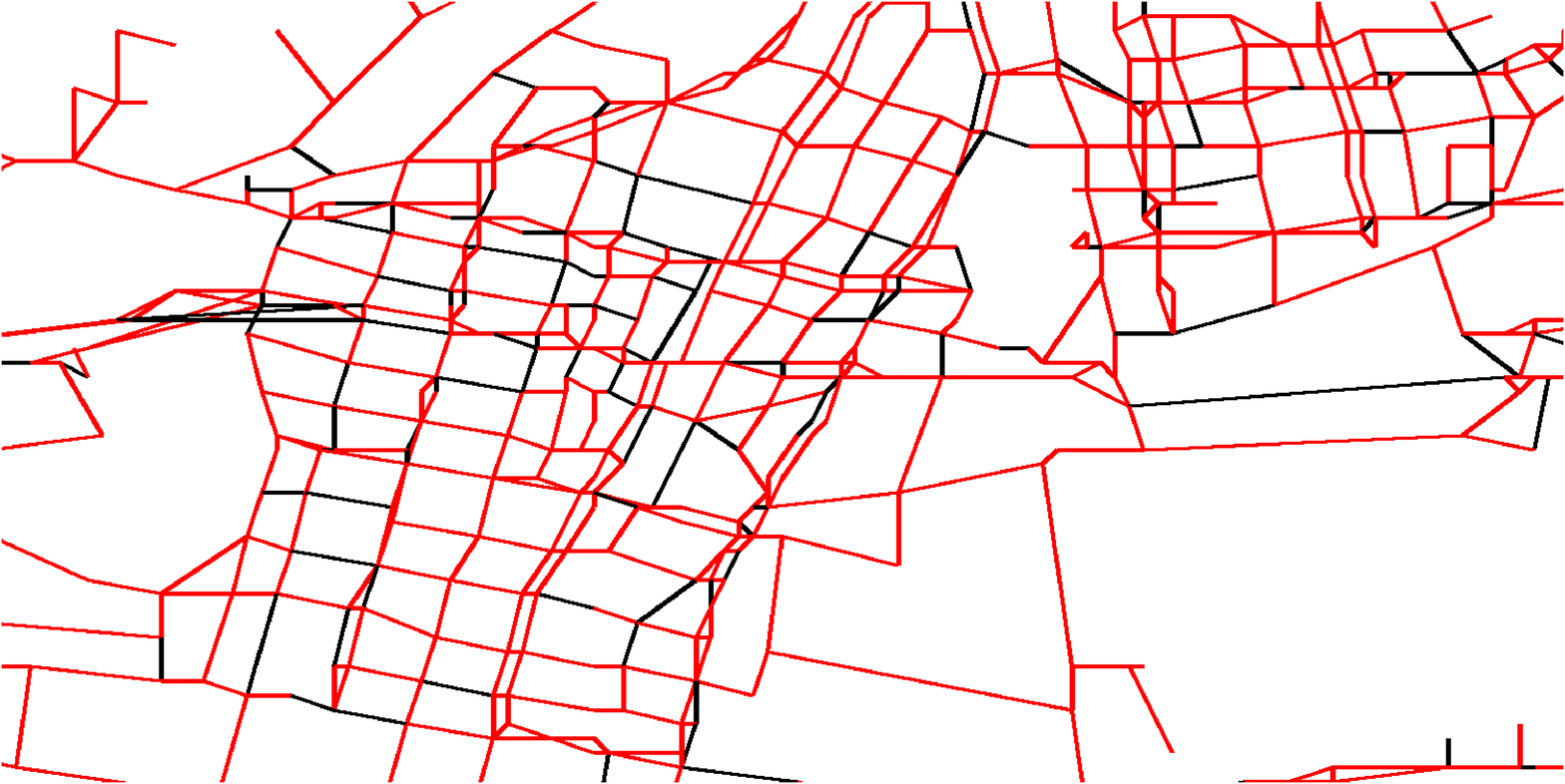} 
\end{center}
\caption{Positions of unobserved roads, where the unobserved roads are colored red.
About 80 \% of roads are unobserved. 
The network in the left panel is the entire network and the network in the right panel is an enlarged image of the center of Sendai.}
\label{fig:lack}
\end{figure}
Figure \ref{fig:lack} shows the positions of the unobserved roads, which are colored red. 
We reconstructed densities in the unobserved roads using the densities in the observed roads, which colored black in the figure.  
Our Bayesian reconstruction result is shown in figure \ref{fig:inf}.
The mean square error (MSE) between the true densities shown in figure \ref{fig:data} and the reconstructed densities shown in figure \ref{fig:inf} 
of the unobserved roads is approximately 0.001447, where the MSE is defined by
\begin{align}
[\mrm{MSE}]:=\frac{1}{\mcal{M}}\sum_{i \in \mcal{M}}\big( x_i^{\mrm{data}} - x_i^{\mrm{recon}}\big)^2, 
\label{eq:MSE}
\end{align}
where $x_i^{\mrm{data}}$ is the true density on road $i$ and $x_i^{\mrm{recon}}$ is the density on road $i$ reconstructed by our method.
The scatter plot of this reconstruction is shown in figure \ref{fig:scatter}.
The correlation coefficient of this scatter plot is approximately 0.919. 
It can be seen that the correlation coefficient is close to one. 
Thus, our simple MRF model can be expected to capture a static statistical property of the traffic data.

\begin{figure}[hbt]
\begin{center}
\includegraphics[height=4.0cm]{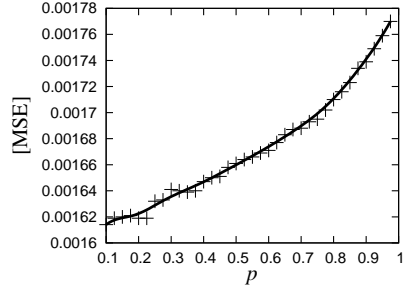}
\end{center}
\caption{MSE versus the missing probability $p$. The solid curve is the Bezier interpolation of points.}
\label{fig:MSE_line_exp}
\end{figure}

Next, we address the average performance of our reconstruction method versus the value of the missing probability $p$.
Figure \ref{fig:MSE_line_exp} shows the MSE versus the value of the missing probability $p$.
Each point is the average value of MSE over 100 trials using the leave-one-out cross-validation method. 
It can be seen that the error increases with the value of the missing probability $p$.

\section{A Statistical Mechanical Analysis of Bayesian Traffic Data Reconstruction} \label{sec:MF-Analysis}

In this section, we clarify the relationship between the model parameters in the MRF model in equation (\ref{eq:MRF_traffic}) 
and the reconstruction performance from a statistical mechanical point of view. 

In our analysis, we assume that the prior probability of traffic data has the same form as our model in equation (\ref{eq:MRF_traffic}), 
\begin{align}
P_{\mrm{prior}}(\bm{x}\mid \bm{\theta})=\frac{1}{Z(\bm{\theta})}\exp\Big(\sum_{i \in V} h_i x_i -\frac{\xi}{2}\sum_{i \in V}x_i^2 
- \frac{J}{2n}\sum_{i<j \in V}(x_i - x_j)^2\Big),
\label{eq:prior-MF}
\end{align}
and assume the values of $\bm{h}$ are independently drawn from the identical distribution $p_h(h)$, 
where notation $Z(\bm{\theta})$ is the normalization constant 
and notation $\sum_{i<j \in V}$ is the summation running over all distinct pairs of vertices $i$ and $j$ in $V = \{1,2,\ldots,n\}$, i.e., 
$\sum_{i<j \in V} = \sum_{i = 1}^n\sum_{j = i + 1}^n$. 
Although road networks have complex structures, 
we neglect the structures and employ the fully-connected model with no structure for the simplicity of analysis.  
For the observations generated from the prior probability in equation (\ref{eq:prior-MF}), we conduct the reconstructions by using the model taking the form
\begin{align}
P_{\mrm{model}}(\bm{x} \mid \bm{\theta}_0)=\frac{1}{Z(\bm{\theta}_0)}\exp\Big(\sum_{i \in V} \beta_i x_i -\frac{\xi_0}{2}\sum_{i \in V}x_i^2 
- \frac{J_0}{2n}\sum_{i<j \in V}(x_i - x_j)^2\Big),
\label{eq:model-MF}
\end{align}
where $\bm{\theta}_0 = \{\bm{\beta}, \xi_0, J_0\}$. 
The bias parameters in the reconstruction model in equation (\ref{eq:model-MF}) are defined by $\beta_i := h_i + \varepsilon_i$, 
and we assume that the values of $\bm{\varepsilon}$ are independently drawn from the identical distribution $p_{\varepsilon}(\varepsilon)$. 
If $J=J_0$, $\xi=\xi_0$, and $\varepsilon_i = 0$, the prior model and the reconstruction model are equivalent.

For the observation with some missing elements, $\bm{x}=\{\bm{x}_{\mcal{M}}, \bm{x}_{\mcal{O}}\}$, generated from the prior model, as shown in equation (\ref{eq:MAP_practical_GGM}), 
the reconstruction using the reconstruction model in equation (\ref{eq:model-MF}) is conducted by 
\begin{align}
x_i^*(\bm{x}_{\mcal{O}}) = \int_{-\infty}^{\infty} x_i P_{\mrm{model}}(\bm{x}_{\mcal{M}} \mid \bm{x}_{\mcal{O}}, \bm{\theta}_0) \diff \bm{x}_{\mcal{M}}, 
\label{eq:MAP_MF}
\end{align}
for $\forall i \in \mcal{M}$.
We measure the statistical performance of the reconstruction by the MSE in equation (\ref{eq:MSE}) averaged over 
all the possible observations generated from the prior probability 
and over all the possible values of bias parameters $\bm{h}$ and $\bm{\varepsilon}$ that are generated from $p_h(h)$ and $p_{\varepsilon}(\varepsilon)$, respectively. 
The averaged MSE is expressed by
\begin{align}
E := \int_{-\infty}^{\infty}\int_{-\infty}^{\infty}\mcal{E}(\bm{h},\bm{\varepsilon})\prod_{i \in V}p_h(h_i)p_{\varepsilon}(\varepsilon_i)
\diff \bm{h} \diff \bm{\varepsilon},
\label{eq:def-aveMSE}
\end{align}
where
\begin{align}
\mcal{E}(\bm{h},\bm{\varepsilon}):=  
\int_{-\infty}^{\infty} P_{\mrm{prior}}(\bm{x}\mid \bm{\theta})\Big(\frac{1}{t}\sum_{i \in \mcal{M}}\big( x_i - x_i^*(\bm{x}_{\mcal{O}}) \big)^2\Big) \diff \bm{x}
\label{eq:def-mcalE}
\end{align}
is the MSE averaged over all the possible observations for the specific biases, where $t := |\mcal{M}|$ is the number of missing elements. 
Equation (\ref{eq:def-aveMSE}) represents the MSE of our Bayesian reconstruction averaged over all the possible situations that appear under our assumption for the prior model.

Since road networks are quite large, we consider the thermal dynamical limit of the mean square error 
by taking limits $n,t \to \infty$ where $p:=t/n$ is fixed at a finite constant. 
Parameter $p$ corresponds to the missing rate; it must be in the interval $[0,1]$.
In the thermal dynamical limit, from equations (\ref{eq:model-MF}) and (\ref{eq:expectation_Conditional_model_limit}), 
we have
\begin{align}
x_i^*(\bm{x}_{\mcal{O}}) = \frac{\beta_i + f(\bm{x}_\mcal{O})}{\xi_0 + J_0}
+\frac{p J_0\big( \mu_h + \mu_{\varepsilon} +f(\bm{x}_{\mcal{O}})\big)}{(\xi_0 + J_0)\big(\xi_0 + (1-p)J_0\big)},
\label{eq:def-xi^*}
\end{align}
where
\begin{align}
f(\bm{x}_{\mcal{O}})= \frac{J_0}{n}\sum_{i \in \mcal{O}}x_i 
\label{eq:def-f0}
\end{align}
(see appendix \ref{app:MFA-BayesianTraffic} for the detailed derivation).
Notations $\mu_h$ and $\mu_{\varepsilon}$ are the averages of $p_h(h)$ and $p_{\varepsilon}(\varepsilon)$, respectively.
From equation (\ref{eq:covarialce_prior_limit}), we have
\begin{align*}
\ave{x_ix_j}_{\mrm{prior}} = 
\begin{cases}
\ave{x_i}_{\mrm{prior}}\ave{x_j}_{\mrm{prior}} & i \not= j\\
1/(\xi + J) +  \ave{x_i}_{\mrm{prior}}^2 & i = j
\end{cases}
,
\end{align*}
where $\ave{\cdots}_{\mrm{prior}}:= \int_{-\infty}^{\infty} (\cdots) P_{\mrm{prior}}(\bm{x}\mid \bm{\theta}) \diff\bm{x}$.
Thus, equation (\ref{eq:def-mcalE}) can be rewritten as
\begin{align*}
\mcal{E}(\bm{h},\bm{\varepsilon})&= \frac{1}{t}\sum_{i \in \mcal{M}} 
\Big( \ave{x_i^2}_{\mrm{prior}} -2\ave{x_i x_i^*(\bm{x}_{\mcal{O}})}_{\mrm{prior}} + \ave{x_i^*(\bm{x}_{\mcal{O}})^2}_{\mrm{prior}}\Big)\nn
&=\frac{1}{t}\sum_{i \in \mcal{M}} 
\Big( \frac{1}{\xi + J} + \ave{x_i}_{\mrm{prior}}^2 -2\ave{x_i}_{\mrm{prior}}\ave{ x_i^*(\bm{x}_{\mcal{O}})}_{\mrm{prior}} 
+ \ave{x_i^*(\bm{x}_{\mcal{O}})^2}_{\mrm{prior}}\Big).
\end{align*}
Similarly, from equation (\ref{eq:covarialce_prior_limit}), 
\begin{align*}
\ave{ f(\bm{x}_{\mcal{O}})^2}_{\mrm{prior}} =\frac{J_0^2}{n^2}\sum_{i \in \mcal{O}}\sum_{j \in \mcal{O}}\ave{x_ix_j}_{\mrm{prior}} 
=\frac{(1-p)J_0^2}{(\xi + J)n} + \Big(\frac{J_0}{n}\sum_{i \in \mcal{O}}\ave{x_i}_{\mrm{prior}}\Big)^2
\underset{n\to \infty}{\longrightarrow } \ave{ f(\bm{x}_{\mcal{O}})}_{\mrm{prior}}^2.
\end{align*}
This means that $f(\bm{x}_{\mcal{O}})$ coincides with $\ave{ f(\bm{x}_{\mcal{O}})}_{\mrm{prior}}$ with a probability of one. 
This leads to relation 
\begin{align*}
\ave{x_i^*(\bm{x}_{\mcal{O}})^2}_{\mrm{prior}} = \ave{x_i^*(\bm{x}_{\mcal{O}})}_{\mrm{prior}}^2,
\end{align*} 
and then, equation (\ref{eq:def-mcalE}) is rewritten as
\begin{align}
\mcal{E}(\bm{h},\bm{\varepsilon})
&=\frac{1}{\xi + J} + \frac{1}{t}\sum_{i \in \mcal{M}} 
 \Big(\ave{x_i}_{\mrm{prior}} -\ave{ x_i^*(\bm{x}_{\mcal{O}})}_{\mrm{prior}}\Big)^2.
\label{eq:mcalE-0}
\end{align}
Since, we find 
\begin{align*}
\ave{f(\bm{x}_{\mcal{O}})}_{\mrm{prior}} \underset{n\to \infty}{\longrightarrow }\frac{(1-p)J_0\mu_h}{\xi} 
\end{align*}
from equation (\ref{eq:expectation_prior_limit}), 
we obtain the average of $x_i^*(\bm{x}_{\mcal{O}})$ in equation (\ref{eq:def-xi^*}) with respect to the prior probability as
\begin{align}
\ave{x_i^*(\bm{x}_{\mcal{O}})}_{\mrm{prior}}
&=\frac{\beta_i }{\xi_0 + J_0} +\frac{J_0  }{\xi(\xi_0 + J_0)}\Big( (1-p) \mu_h
+\frac{p \big(\xi (\mu_h + \mu_{\varepsilon}) +(1-p)J_0\mu_h \big)}{\xi_0 + (1-p )J_0\mu_h}\Big).
\label{eq:ave-xi^*}
\end{align}
By using equations (\ref{eq:def-aveMSE}), (\ref{eq:mcalE-0}), (\ref{eq:ave-xi^*}), and (\ref{eq:expectation_prior_limit}), 
with a straightforward calculation, we finally obtain the explicit form of equation (\ref{eq:def-aveMSE}) as
\begin{align}
E  = \frac{1}{\xi + J} + \Big( \frac{\sigma_h}{\xi+J} - \frac{\sigma_h}{\xi_0 + J_0}\Big)^2+  \frac{\sigma_{\varepsilon}^2 }{(\xi_0 + J_0)^2}
+\bigg(\frac{ (\xi - \xi_0)\mu_h + \xi \mu_{\varepsilon}}{\xi\big(\xi_0 + (1-p )J_0\big)}\bigg)^2,
\label{eq:aveMSE}
\end{align}
where $\sigma_h^2$ and $\sigma_{\varepsilon}^2$ denote the variances of $p_h(h)$ and $p_{\varepsilon}(\varepsilon)$, respectively. 
Equation (\ref{eq:aveMSE}) does not require the information about which roads are selected as the unobserved. 
The dependency on these data disappears as a result of the averaging operations. 

It is obvious that $E$ takes the minimum value 
\begin{align}
E_{\mrm{min}} = \frac{1}{\xi + J}
\end{align}
when there is no model error between the prior model in equation (\ref{eq:prior-MF}) and the reconstruction model in equation (\ref{eq:model-MF}), that is, 
when $J=J_0$, $\xi=\xi_0$, and $\mu_{\varepsilon} = \sigma_{\varepsilon}= 0$. 
In the following, we examine numerically the relation between the averaged MSE and the parameters in the reconstruction model. 
For the numerical experiments, we set the parameters in the prior model to $J = 1$, $\xi = 0.2$, $\mu_h = 1$, and $\sigma_h = 0.5$.

\begin{figure}[hbt]
\begin{center}
\includegraphics[height=4.5cm]{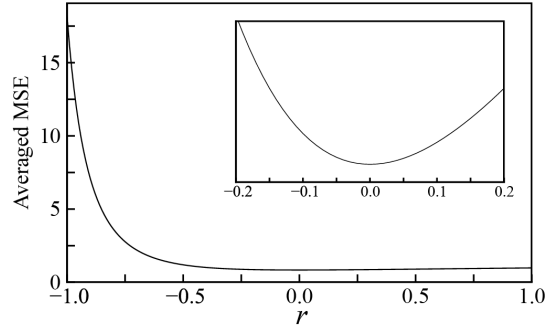}
\end{center}
\caption{Plot of $E$ against error $r$ when $\xi_0=\xi$, $\mu_{\varepsilon} = \sigma_{\varepsilon}= 0$, $p = 0.5$, and $J_0 = J + r$.
The vertical axis is $E$ in equation (\ref{eq:aveMSE}). The inset is an enlarged plot around $r=0$. }
\label{fig:ErJ}
\end{figure}

First, we examine the relation when a model error exists only in the interaction parameters, that is, 
when $\xi_0=\xi$, $\mu_{\varepsilon} = \sigma_{\varepsilon}= 0$, and $J_0 = J + r$. The parameter $r$ is the error of interaction.
Figure \ref{fig:ErJ} shows the plot of $E$ against error $r$ when $p = 0.5$.
It can be seen that, where $J_0$ is bigger than $J$, the performance level is relatively robust. 
In contrast, the performance level drastically decreases when $J_0$ is smaller than $J$.

From equation (\ref{eq:aveMSE}), it can be seen that the dependency on the missing rate $p$ arises when model errors exist in the bias parameters 
or in variance parameters $\xi$ and $\xi_0$. 
Next, we consider the case where model errors exist in the bias parameters and the variance parameters.
\begin{figure}[hbt]
\begin{center}
\includegraphics[height=4.5cm]{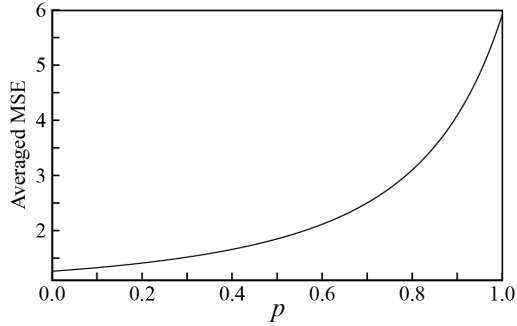}
\end{center}
\caption{Plot of $E$ against missing rate $p$ when $\xi_0=0.4$, $\mu_{\varepsilon}=0.1$, $\sigma_{\varepsilon}= 0$, and $J_0 = J$.
The vertical axis is $E$ in equation (\ref{eq:aveMSE}). }
\label{fig:ErBias}
\end{figure}
Figure \ref{fig:ErBias} shows the plot of $E$ against missing rate $p$ when $\xi_0=0.4$, $\mu_{\varepsilon}=0.1$, $\sigma_{\varepsilon}= 0$, and $J_0 = J$. 
The reconstruction performance level decreases with the increase in the value of $p$. 
This performance behavior seems to be qualitatively similar to the behavior of our numerical traffic density reconstruction in figure \ref{fig:MSE_line_exp}.  
From this result, we can presume that the behavior of MSE in figure \ref{fig:MSE_line_exp} is caused primarily by the model errors in either the biases or the variance, or both. 

The parameters in our reconstruction model were determined by the training data as described in section \ref{sec:Numerical-Reconstruction}. 
However, in the training, it was not easy to find the truly optimal values of parameters from the training data,  
because the number of training data was much smaller than the number of parameters. 
This could be one of the reasons why the model errors exist.

\section{Conclusion} \label{sec:conclusion}

In this paper, we introduced the Bayesian reconstruction framework, in which missing data are probabilistically interpolated, 
and overviewed the application of Bayesian reconstruction to the problem of traffic data reconstruction in the field of traffic engineering. 
Our traffic reconstruction model in equation (\ref{eq:MRF_traffic}) neglects some real traffic properties, for example, traffic lanes, contraflows, and so on.
Nevertheless, the results of our reconstruction seem to be accurate. 
It can be expected that our simple GGM captures the static statistical property of traffic data 
and that it can be used an important base model of Bayesian traffic reconstructions. 
The extension of our model by taking real traffic properties into account should be addressed in the next study. 

In the latter part of this paper, 
we evaluated the statistical performance of our reconstruction by using the statistical mechanical analysis, that is mean-field analysis. 
In our analysis, we used the simplified reconstruction model, which has no network structures, for the convenience of calculations. 
However, since real road networks are network structures, the result of our evaluation can be only a rough approximation. 
The authors proposed a method based on the belief propagation method 
to evaluate the statistical properties of Bayesian reconstructions on structured networks in the context of image processing~\cite{KYK2012}. 
We strongly suggest that the method can be applied to Bayesian traffic reconstruction and can lead to more realistic evaluations.

\appendix

\section{Mean-field Analysis for Traffic Data Reconstruction Model} \label{app:MFA-BayesianTraffic}

The partition function in equation (\ref{eq:prior-MF}) is
\begin{align*}
Z(\bm{\theta})&= \int_{-\infty}^{\infty}\exp\Big(\sum_{i \in V} h_i x_i -\frac{\xi}{2}\sum_{i \in V}x_i^2 
- \frac{J}{2n}\sum_{i<j \in V}(x_i - x_j)^2\Big) \diff\bm{x}\nn
&=\int_{-\infty}^{\infty}\exp\Big(\sum_{i \in V} h_i x_i -\frac{1}{2}\Big( \xi + \frac{(n-1)J}{n} \Big)\sum_{i \in V}x_i^2 
+ \frac{J}{2}\sum_{i<j \in V}x_jx_j\Big) \diff\bm{x}.
\end{align*}
By using the Hubbard-Stratonovich transformation, we have
\begin{align*}
Z(\bm{\theta})
&=\int_{-\infty}^{\infty}\exp\Big(\sum_{i \in V} h_i x_i -\frac{\xi + J}{2}\sum_{i \in V}x_i^2 
+ \frac{Jn}{2}\Big(\frac{1}{n} \sum_{i \in V}x_i \Big)^2\Big) \diff\bm{x}\nn
&=\sqrt{\frac{Jn}{2\pi}}\int_{-\infty}^{\infty}\int_{-\infty}^{\infty}
\exp\Big(\sum_{i \in V} \big(h_i + z J\big) x_i -\frac{\xi + J}{2}\sum_{i \in V}x_i^2  - \frac{Jn}{2}z^2\Big) \diff\bm{x}\diff z.
\end{align*}
The Gaussian integral leads to
\begin{align}
Z(\bm{\theta}) = \sqrt{\frac{\xi + J}{\xi}}\Big( \frac{2\pi}{\xi + J}\Big)^{n/2} 
\exp \Big(\frac{1}{2n(\xi + J)}\sum_{i \in V} h_i^2 +\frac{J}{2n^2 (\xi + J)\xi}\Big(\sum_{i \in V} h_i \Big)^2 \Big).
\end{align}
Therefore, the free energy (per one variable) for the prior model in (\ref{eq:prior-MF}) is expressed by
\begin{align}
F(\bm{\theta}):=-\frac{1}{n} \ln Z(\bm{\theta})
=-\frac{1}{2n}\ln \frac{\xi + J}{\xi} - \frac{1}{2}\ln \frac{2\pi}{\xi + J}
-\frac{1}{2n(\xi + J)}\sum_{i \in V} h_i^2 -\frac{J}{2n^2 (\xi + J)\xi}\Big(\sum_{i \in V} h_i \Big)^2, 
\label{eq:FreeEnergy-Prior}
\end{align}
so that the first- and the second-order moments of the prior model are given by
\begin{align}
\ave{x_i}_{\mrm{prior}}= - n \frac{\partial F(\bm{\theta})}{\partial h_i}=\frac{h_i}{\xi + J}+ \frac{J}{n (\xi + J)\xi}\sum_{j\in V} h_j 
\label{eq:expectation_prior}
\end{align}
and
\begin{align}
\ave{x_i x_j}_{\mrm{prior}} 
= - n \frac{\partial^2 F(\bm{\theta})}{\partial h_i\partial h_j} + \ave{x_i}_{\mrm{prior}}\ave{x_j}_{\mrm{prior}}
=\frac{\delta_{ij}}{\xi + J} + \frac{J}{n (\xi + J)\xi}+ \ave{x_i}_{\mrm{prior}}\ave{x_j}_{\mrm{prior}},
\label{eq:covariance_prior}
\end{align}
respectively, 
where $\ave{\cdots}_{\mrm{prior}}= \int_{-\infty}^{\infty} (\cdots) P_{\mrm{prior}}(\bm{x}\mid \bm{\theta}) \diff\bm{x}$ and 
$\delta_{ij}$ is Kronecker's delta.

Next, we find the free energy for the conditional probability of the reconstruction model in equation (\ref{eq:MAP_MF}). 
The conditional probability is expressed by
\begin{align*}
P_{\mrm{model}}(\bm{x}_{\mcal{M}} \mid \bm{x}_{\mcal{O}}, \bm{\theta}_0)
=\frac{1}{\mcal{Z}(\bm{\theta}_0,\bm{x}_{\mcal{O}})}\exp\Big(\sum_{i \in \mcal{M}} \big(\beta_i  + f(\bm{x}_{\mcal{O}})\big)x_i 
-\frac{1}{2}\Big( \xi_0 + \frac{(n-1)J_0}{n} \Big)\sum_{i \in \mcal{M}}x_i^2 + \frac{J_0}{n}\sum_{i<j \in \mcal{M}}x_ix_j\Big),
\end{align*}
where $f(\bm{x}_{\mcal{O}})$ is defined in equation (\ref{eq:def-f0}) 
and $\sum_{i<j \in \mcal{M}}$ represents the summation running over all distinct pairs of vertices $i$ and $j$ in $\mcal{M} \subseteq V$.
$\mcal{Z}(\bm{\theta}_0,\bm{x}_{\mcal{O}})$ represents the partition function defined by
\begin{align*}
\mcal{Z}(\bm{\theta}_0,\bm{x}_{\mcal{O}}):=\int_{-\infty}^{\infty} \exp\Big(\sum_{i \in \mcal{M}} \big(\beta_i  + f(\bm{x}_{\mcal{O}})\big)x_i 
-\frac{1}{2}\Big( \xi_0 + \frac{(n-1)J_0}{n} \Big)\sum_{i \in \mcal{M}}x_i^2 + \frac{J_0}{n}\sum_{i<j \in \mcal{M}}x_ix_j\Big)\diff \bm{x}_{\mcal{M}}.
\end{align*}
Using almost the same derivation as the above mean-field derivation for the prior model, we find
\begin{align}
\mcal{Z}(\bm{\theta}_0,\bm{x}_{\mcal{O}})&=\sqrt{\frac{\xi_0 + J_0}{\xi_0 + (1-p )J_0}} \Big(\frac{2\pi}{\xi_0 + J_0}\Big)^{t/2}
\exp\Big\{\frac{1}{2(\xi_0 + J_0)}\sum_{i \in \mcal{M}}\big(\beta_i + f(\bm{x}_\mcal{O}) \big)^2\nn
\aleq+\frac{p J_0}{2t (\xi_0 + J_0)\big(\xi_0 + (1-p )J_0\big)}\Big(\sum_{i\in \mcal{M}}\big(\beta_i +f(\bm{x}_\mcal{O}) \big)\Big)^2 \Big\},
\label{eq:Z-conditional_model}
\end{align}
where $t$ is the number of missing elements defined in section \ref{sec:MF-Analysis} 
and $p = t /n$ is associated with the missing rate. 
From equation (\ref{eq:Z-conditional_model}), we can obtain the free energy (per one variable) for the conditional probability of the reconstruction model by
\begin{align}
\mcal{F}(\bm{\theta}_0,\bm{x}_{\mcal{O}}):=-\frac{1}{t}\ln \mcal{Z}(\bm{\theta}_0,\bm{x}_{\mcal{O}})
&=-\frac{1}{2t} \frac{\xi_0 + J_0}{\xi_0 + (1-p )J_0}- \frac{1}{2}\ln \frac{2\pi}{\xi_0 + J_0}
-\frac{1}{2t(\xi_0 + J_0)}\sum_{i\in \mcal{M}}\big(\beta_i + f(\bm{x}_{\mcal{O}}) \big)^2\nn
\aleq
-\frac{p J_0}{2t^2 (\xi_0 + J_0)\big(\xi_0 + (1-p )J_0\big)}\Big(\sum_{i\in \mcal{M}}\big(\beta_i +f(\bm{x}_\mcal{O}) \big)\Big)^2, 
\label{eq:FreeEnergy-Conditional_model}
\end{align}
so that the first-order moments of the conditional probability of the reconstruction model are given by
\begin{align}
\ave{x_i}_{\mcal{M}\mid \mcal{O}}=- t \frac{\partial \mcal{F}(\bm{\theta}_0,\bm{x}_{\mcal{O}})}{\partial \beta_i}=\frac{\beta_i + f(\bm{x}_\mcal{O})}{\xi_0 + J_0}
+\frac{p J_0}{t (\xi_0 + J_0)\big(\xi_0 + (1-p)J_0\big)}\sum_{j\in \mcal{M}}\big(\beta_j +f(\bm{x}_{\mcal{O}}) \big)
\label{eq:expectation_Conditional_model}
\end{align}
for $\forall i \in \mcal{M}$, where we define 
$\ave{\cdots}_{\mcal{M}\mid \mcal{O}}:= \int_{-\infty}^{\infty} (\cdots) P_{\mrm{model}}(\bm{x}_{\mcal{M}} \mid \bm{x}_{\mcal{O}}, \bm{\theta}_0)\diff\bm{x}$. 

We consider the thermal dynamical limit by taking limits $n,t \to \infty$, where $p$ is fixed at a finite constant. 
In the thermal dynamical limit, the free energies in equations (\ref{eq:FreeEnergy-Prior}) and (\ref{eq:FreeEnergy-Conditional_model}) are reduced to
\begin{align*}
F(\bm{\theta})= -\frac{1}{2}\ln \frac{2\pi}{\xi + J}
-\frac{\sigma_h^2 + \mu_h^2}{2(\xi + J)} -\frac{J \mu_h^2}{2 (\xi + J)\xi}
\end{align*}
and
\begin{align*}
\mcal{F}(\bm{\theta}_0,\bm{x}_{\mcal{O}})=-\frac{1}{2}\ln \frac{2\pi}{\xi_0 + J_0}
-\frac{\sigma_h^2 + \sigma_{\varepsilon}^2 + \big(\mu_h + \mu_{\varepsilon} +  f(\bm{x}_\mcal{O})\big)^2 }{2(\xi_0 + J_0)}
-\frac{p J_0 \big(\mu_h + \mu_{\varepsilon} +f(\bm{x}_\mcal{O}) \big)^2}{2 (\xi_0 + J_0)\big(\xi_0 + (1-p )J_0\big)}, 
\end{align*}
respectively, where $\mu_h$ and $\sigma_h^2$ are the average and variance of $p_h(h)$, respectively, 
and $\mu_{\varepsilon}$ and $\sigma_{\varepsilon}^2$ are the average and variance of $p_{\varepsilon}(\varepsilon)$, respectively. 
$\{\mu_h, \mu_{\varepsilon}, \sigma_h^2, \sigma_{\varepsilon}^2\}$ appear because of the law of large numbers. 
Similarly, the moments in equations (\ref{eq:expectation_prior}), (\ref{eq:covariance_prior}), and (\ref{eq:expectation_Conditional_model}) are reduced to
\begin{align}
\ave{x_i}_{\mrm{prior}}=\frac{h_i}{\xi + J}+ \frac{J \mu_h}{(\xi + J)\xi},  
\label{eq:expectation_prior_limit}\\
\ave{x_i x_j}_{\mrm{prior}} 
=\frac{\delta_{ij}}{\xi + J} + \ave{x_i}_{\mrm{prior}}\ave{x_j}_{\mrm{prior}},
\label{eq:covarialce_prior_limit}
\end{align}
and
\begin{align}
\ave{x_i}_{\mcal{M}\mid \mcal{O}}=\frac{\beta_i + f(\bm{x}_\mcal{O})}{\xi_0 + J_0}
+\frac{p J_0\big( \mu_h + \mu_{\varepsilon} +f(\bm{x}_{\mcal{O}})\big)}{(\xi_0 + J_0)\big(\xi_0 + (1-p)J_0\big)},
\label{eq:expectation_Conditional_model_limit}
\end{align}
respectively, in the thermal dynamical limit.

\subsubsection*{Acknowledgment}
The authors are very grateful to Professor Masao Kuwahara and Dr. Jinyoung Kim of the Graduate School of
Information Science, Tohoku University for providing road network data and traffic simulation data. 
This work was partially supported by grants-in-aid (nos. 25280089, 24700220, and 25-7259) from the Ministry of Education, Culture, Sports, Science and Technology of Japan.

%\bibliographystyle{splncs}
%\bibliography{egbib}

\end{document}